\documentclass[10pt,twocolumn,letterpaper]{article}

\usepackage{iccv}
\usepackage{times}
\usepackage{epsfig}
\usepackage{graphicx}
\usepackage{amsmath}
\usepackage{amssymb}
\usepackage{multirow}
\usepackage{algorithm}
\usepackage{algorithmic}
\usepackage{float}
\usepackage{subfigure}
\usepackage[font={small}]{caption}

\usepackage[pagebackref=true,breaklinks=true,letterpaper=true,colorlinks,bookmarks=false]{hyperref}

\iccvfinalcopy % *** Uncomment this line for the final submission

\def\iccvPaperID{1025} % *** Enter the ICCV Paper ID here

\ificcvfinal\fi
\begin{document}

%%%%%%%%% TITLE
\title{Semi-Supervised Image-to-Image Translation using Latent Space Mapping}

\author{Pan Zhang$^1$
	\thanks{Author did this work during the internship at Microsoft Research Asia.}
	, Jianmin Bao$^2$, Ting Zhang$^2$, Dong Chen$^2$, Fang Wen$^2$
	\\
	$^1$University of Science and Technology of China \quad
	$^2$Microsoft Research Asia
}

% \and

\twocolumn[{%
\renewcommand\twocolumn[1][]{#1}%
\vspace{-1em}
\maketitle
\vspace{-1em}
\begin{center}
    \centering
    \includegraphics[width=\linewidth]{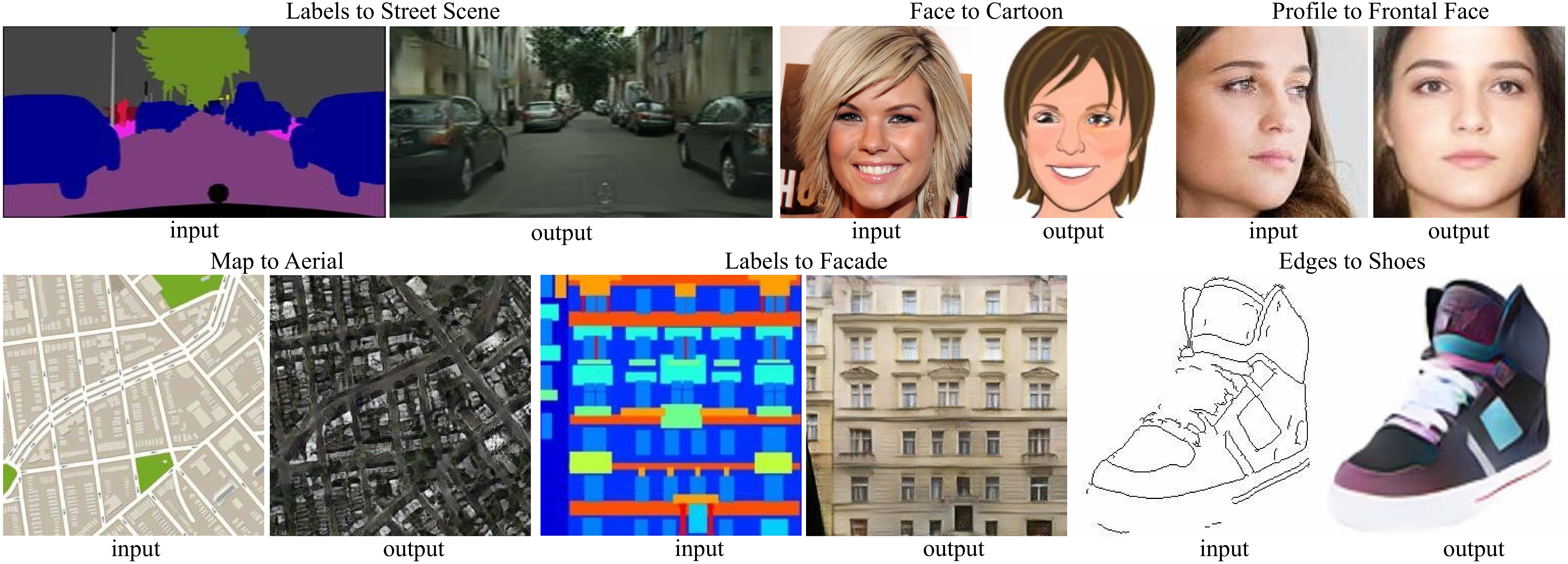}
    \captionof{figure}{We introduce a novel framework for semi-supervised image translation which can be applied to various image-to-image translation tasks. The key idea is to apply the transformation over the image features instead of the original images, resulting in improved quality of translation results even using a small number of paired data. Here we show results of the proposed method on several examples.}
    \vspace{0.4cm}
    \label{fig:teaser}
\end{center}%
}]

\maketitle
%\thispagestyle{empty}

%%%%%%%%% ABSTRACT
\begin{abstract}
	Recent image-to-image translation works have been transferred from supervised to unsupervised settings due to the expensive cost of capturing or labeling large amounts of paired data. 
	However, current unsupervised methods using the cycle-consistency constraint may not find the desired mapping, especially for difficult translation tasks. 
	On the other hand, a small number of paired data are usually accessible.
	We therefore introduce a general framework for semi-supervised image translation. 
	Unlike previous works, our main idea is to learn the translation over the latent feature space instead of the image space.
	Thanks to the low dimensional feature space, it is 
	easier to find the desired mapping function,
	resulting in improved quality of translation results
	as well as the stability of the translation model.
	Empirically we show that using feature translation generates better results, even using a few bits of paired data.
	Experimental comparisons with state-of-the-art approaches demonstrate the effectiveness of the proposed framework on a variety of challenging image-to-image translation tasks\footnote{This paper finished in March 2019.}.
\end{abstract}
\vspace{-0.15in}

%%%%%%%%% BODY TEXT
\section{Introduction}
Image-to-image translation aims to learn a mapping that transforms an input image in the source domain into a corresponding output image in the target domain, which can be applied to a wide range of applications, \eg, image colorization~\cite{isola2017image}, semantic labels to real image~\cite{isola2017image}, photo cartoonization~\cite{wen2013cartoon, chen2018cartoongan}, style transfer~\cite{gatys2016image}, attribute transfer~\cite{choi2018stargan}, and sketch-to-image generation~\cite{lu2018image}.

If there is enough paired data from both domains, a simple encoder-decoder network is sufficient to obtain satisfactory results~\cite{isola2017image,wang2018high,xian2018texturegan,lu2018image,song2018geometry,dekel2017smart}. The encoder extracts the relevant information from the input domain and passes it on to the decoder,
which then decodes this information to generate an image from the output domain. 
However, large amount of supervised data can be difficult and expensive to obtain. Many recent works~\cite{zhu2017unpaired,kim2017learning,yi2017dualgan,liu2017unsupervised} have proposed conducting training without any paired data.

Such unsupervised methods suffer from instability during training, since there exist infinite possible mapping functions satisfying the cycle-consistency constraint which reconstructs the input image by mapping back the translated input image.
This undermines the generality of the unsupervised methods over difficult translation tasks, \eg, 
the translation of profile faces to frontal faces (face frontalization) which involves geometric shape changing.

We observe that it is always possible to get some paired data in some situations.
For example, for face frontalization, although it is hard to acquire a person's frontal face and profile face at the same time, the Multi-PIE database~\cite{gross2010multi} still contains a few such paired data.
It is crucial to make good use of this existing knowledge to guide the model, 
especially when the amount of the pairs is small.
Therefore, the interest of this paper is in semi-supervised image-to-image translation.

Unlike previous works which directly transform the input image to the target image, we propose performing translation in the latent space instead of the image space.
The lower-dimensional latent space allows
the translation model easier to find the desired mapping, leading to high quality image generation results with less paired images. 
Empirically we show that 1) feature translation improves the quality of the generated images under various settings;
and 2) feature translation is able to achieve better results with only a small number of paired data than the image translation with full supervision.
Besides higher quality generation results,
another advantage of our approach compared with unsupervised image translation works,
which adopt the cycle-consistency constraint,
is that our model takes much less training time.

Our framework is a general-purpose solution for image translation tasks and can be applied to a wide variety of applications,
including profile to frontal faces, semantic labels to street photos, edge maps to photographs,
real faces to cartoons and so on, which is shown in Figure~\ref{fig:teaser}. 
We also compare our model against state-of-the-art approaches to show that our framework produces superior generation results.

\section{Related Work}

\noindent\textbf{Image-to-Image Translation.} 
The goal of image-to-image translation is to convincingly translate an input image from one domain to another.
With the success of generative techniques in image generation, promising progress has been made in image-to-image translation.
Existing approaches can be roughly divided into three categories: supervised, unsupervised and semi-supervised.

Supervised translation works aim to learn the translation model using the provided matching image pairs.
Pix2pix~\cite{isola2017image} proposes a unified framework for image-to-image translation based on conditional GANs, which has been extended to generating high-resolution images in pix2pixHD~\cite{wang2018high}.
There exist other efforts~\cite{xian2018texturegan,lu2018image,song2018geometry,dekel2017smart}
focusing on specific applications, \eg, transforming sketches, contours, or hand sketches to images. 
They usually use $L_1$ loss~\cite{huang2017beyond} , adversarial loss~\cite{isola2017image}, or perceptual loss~\cite{johnson2016perceptual} at the image level. 
However, it is well-known that a large amount of supervised data is expensive and difficult to obtain.

Many recent works~\cite{hoshen2018nam,hoshen2018non,zhu2017unpaired,liu2017unsupervised} have been devoted to unsupervised image translation.
The most popular constraint is cycle-consistency: reconstructing the input sample by
translating back the translated input result, which is adopted in CycleGAN~\cite{zhu2017unpaired}, DiscoGAN~\cite{kim2017learning}, DualGAN~\cite{yi2017dualgan}, and UNIT~\cite{liu2017unsupervised}. 
However, these unsupervised methods are still inferior to supervised methods
and are not able to handle complex translation tasks, since there is no paired data showing how an image is transformed from one domain to another domain.

Another class is semi-supervised translation works, such as ALICE~\cite{li2017alice}, Triple GAN~\cite{chongxuan2017triple} and $\Delta$-GAN~\cite{gan2017triangle}.
These models explore supervision 
from a few paired samples to guide the model which is unstable in the unsupervised setting.
Nonetheless, in image translation task, they are still not comparable with supervised models.
Our work also belongs to the semi-supervised category.
We propose learning the translation in the latent space instead of the image space.
We show that using the proposed feature translation, our model given only a few
supervision is able to achieve better performance than the supervised method with full supervision.

\vspace{0.1cm}
\noindent\textbf{Latent Space Constraint.} 
Recent works have also explored the latent space by 1) introducing the feature cycle-consistency, \ie, 
minimizing the distance between the latent features of the real images 
and the generated images, such as 
the $f$-constancy term introduced in DTN~\cite{taigman2016unsupervised}, the latent reconstruction loss in MUNIT~\cite{Huang_2018_ECCV} and the latent cycle-consistency in XNet~\cite{sendik2019xnet};
2) assuming a shared latent (content) space so that the latent (content) features of the corresponding paired images are the same,
\eg, UNIT~\cite{liu2017unsupervised} and MUNIT~\cite{Huang_2018_ECCV}.
In contrast to the above prior works which introduce latent space constraint implicitly over the image level,
the key idea of ours is to explicitly learn the translation model over the feature space, whose dimension is much lower than the image space, leading to the ease of model training and better translations.

\section{Latent Space Translation Framework}
In this section, we formally introduce our method.
In the problem of semi-supervised image-to-image translation, we have two sets of
images from two domains respectively, \ie, $\mathcal{X}_1$ and $\mathcal{X}_2$,
in which there is a partial (\eg, $\approx 10\%$) paired dataset $\mathcal{P}=\{p_1, p_2\}$ with $p_1\in\mathcal{X}_1$ and
$p_2\in\mathcal{X}_2$.
The goal is to learn an image-to-image translation model that can faithfully transform images from the source domain to the target domain.

\begin{figure}[t]
    \centering
    \includegraphics[width=1.0\linewidth]{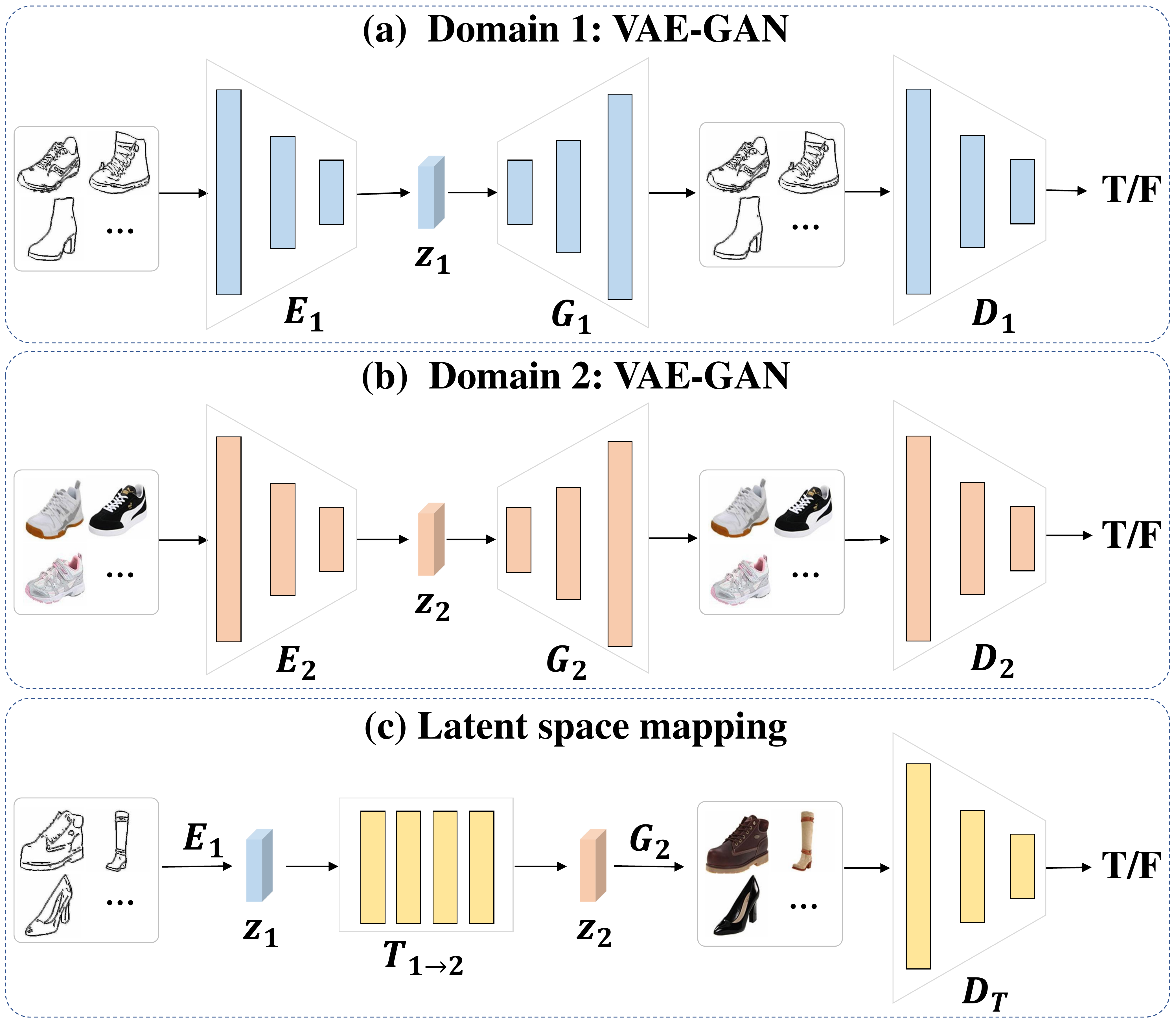}
    \vspace{-0.6cm}
    \caption{Illustrating the proposed framework. The losses are omitted for clarity. (a) We first train a VAE-GAN model using all the images in the first domain. (b) Then we train another VAE-GAN model using all the images in the second domain. (c) At last, we learn the latent space mapping using the paired feature. In the test process, we use $E_1$, $T_{1\to 2}$, and $G_2$ for image translation from domain $1$ to domain $2$.}
    \label{fig:framework}
    \vspace{-0.5cm}
\end{figure}

Towards this goal, we introduce our latent space translation framework. Figure~\ref{fig:framework} shows an overview of our framework, which can be divided into two parts: 1) two auto-encoder GANs are independently learned in each domain, mapping the images to the latent feature space; and 2) a transfer network learns the translation between the latent spaces of the two domains. Next we will introduce the details of these two parts.

\vspace{0.1cm}
\noindent
{\bf VAE-GAN.}
As illustrated in the figure, for each domain, we adopt the variational auto-encoder GAN (VAE-GAN)~\cite{larsen2015autoencoding} to extract the latent feature. The VAE-GAN combines VAEs and GANs to simultaneously learn the encoder, decoder and discriminator for generating high-quality images. 

Without loss of generality, we use the first domain as an example.
A VAE contains two neural networks: an encoder $E_1$ and a decoder $G_1$.
Given a data sample $x_1 \in \mathcal{X}_1$, the encoder maps the data sample to 
a latent representation code $z_1$ and then the decoder maps the latent code back
to the data $\bar{x}_1$, \ie,
{\small
 \begin{align}
 \bar{x}_1  \sim p(x_1|z_1)= G_1 (z_1), ~ z_1 \sim q(z_1|x_1)=E_1(x_1).
 \end{align}
}
The VAE assumes that the variational approximate posterior $q(z_1|x_1)$ is a multivariate Gaussian with a diagonal covariance structure. In our formulation, we choose the identity matrix as the covariance matrix. 
We adopt the reparameterization trick in~\cite{KingmaW13} to let the sampling operation be differentiable and hence enable back-propagation in networks.

A GAN also contains two neural networks: a generator $G_1$ and a discriminator $D_1$.
The generator maps a latent code to an image, and the discriminator outputs 
a probability that the generated image is real.
The goal of the generator is to produce realistic images that can fool the discriminator,
whereas the discriminator tries to distinguish the generated images from real images as much as possible.
The two networks play a zero-sum game until a equilibrium state is achieved, where the generated images are
indistinguishable from real images.

The VAE loss aims to minimize the variational upper bound that consists of the regularization term over the encoder and the minus expected reconstruction error, \ie,
\begin{equation}
\mathcal{L}_{VAE_1} =  D_{KL}(q(z_1|x_1) || p(z_1)) - \mathbb{E}_{q(z_1|x_1)}[\text{log}~ p(x_1|z_1)],
%\mathcal{L}_{VAE_v} =  D_{KL}(q(z_v|x_v) || p(z_v)) - \mathbb{E}_{q(z_v|x_v)}[\text{log}~ p(x_v|z_v)],
\end{equation}
where $D_{KL}$ is the Kullback-Leibler divergence. 
The widely used adversarial loss for a GAN is 
$\mathbb{E}[\text{log}~D_1(x_1)] + \mathbb{E}[\text{log}~(1-D_1(G_1(z_1)))]$.
Here we use LSGANs~\cite{mao2017least} for stable training, that is,
\begin{equation}
\mathcal{L}_{GAN_1} = \mathbb{E}[D_1(x_1)^2] + \mathbb{E}[(1-D_1(G_1(z_1)))^2].
%\mathcal{L}_{GAN_v} = \mathbb{E}[\text{log}~D_v(x_v)] + \mathbb{E}[\text{log}~(1-D_v(G_v(z_v)))].
\end{equation}
Similarly, we can get the loss $\mathcal{L}_{VAE_2}$ and $\mathcal{L}_{GAN_2}$ for the second domain in the same way.

\vspace{0.1cm}
\noindent
{\bf Feature Transformation.}
Given the paired dataset $\mathcal{P}$, 
previous works usually learn the translation model directly on the image level,
\ie, transforming $p_1$ to $p_2$ through the learned model.
Consider the dimension of the image space, which is usually hundreds of thousands (\eg, $196,608=256\times 256\times 3$),
it may be easier to find the desired translation function if the transformed space could be low-dimensional.
Inspired from this,
our work is interested in the translation on the low-dimension latent feature level.

Specifically, after training the VAE-GANs, the paired image dataset can be converted to the corresponding paired latent feature dataset
by applying the encoders $E_1$ and $E_2$ over the images.
Let $\mathcal{F} = \{z_1,z_2\}$ denote the paired feature dataset with $z_1 = E_1(p_1)$ and $z_2 = E_2(p_2)$.
We learn a feature translation model $T_{1\to 2}: z_1 \to z_2$
with two objectives:
1) the transformed feature should be close to the target feature as much as possible;
2) the generated images from the transformed features should be indistinguishable from the real images.

Formally, the loss function for learning the feature translation model consists of two terms:
one is $L_1$ loss penalizing the deviation of the transformed feature from the target feature;
the other is the adversarial loss that encourages the generated images to look natural.
The loss function is written as,
\begin{align}
\mathcal{L}_{FT_{1}} =  \mathcal{L}_{adv_1} + &\lambda_f \mathcal{L}_{T_{1\to 2}}, \label{eqn:translation}
\end{align}
and $\mathcal{L}_{adv_1}=\mathbb{E}[D_T(p_2)^2]  
 + \mathbb{E}[(1-D_T(G_2(T_{1\to 2}(z_1))))^2]$,
where $D_T$ is the discriminator for the feature translation model.
$\lambda_f$ is a weight controlling the balance of the two loss terms, and
$\mathcal{L}_{T_{1\to 2}}=\|T_{1\to 2}(z_1) - z_2\|_1$.

\vspace{0.1cm}
\noindent
{\bf Feature Matching Loss.} 
We adopt feature matching loss in order to improve GAN loss,
making GAN training more stable and the generated results more realistic.
As with pix2pixHD~\cite{wang2018high},
we match the intermediate representations of the real images and the synthesized images
using the intermediate outputs of the discriminator at multiple scales. We also add a perceptual loss based on several intermediate layers of the VGG network.
The feature matching loss is given as,
\begin{align}
\mathcal{L}_{FM}(x,\bar{x}) 
 =& \sum_i \frac 1 {M_{GAN}^i}\| F_{GAN}^i (x) - F_{GAN}^i (\bar{x})\|_1 \nonumber \\
+ & \sum_i \frac 1 {M_{VGG}^i} \| F_{VGG}^i (x) - F_{VGG}^i (\bar{x})\|_1,
\end{align}
where $F_{GAN}^i$ ($F_{VGG}^i$) denotes the $i$th intermediate feature of the GAN discriminator (VGG network), and $M_{GAN}^i$ ($M_{VGG}^i$) indicates the total number of elements in that feature map.

\vspace{0.1cm}
\noindent
{\bf Training Procedure.} 
Our framework adopts a two-stage training strategy: we first train a VAE-GAN for each domain,
and then train the translation network with the learned VAE-GAN fixed.
Specifically, the loss function of training the VAE-GAN for the first domain is,
\begin{align}
\min_{E_1,G_1}\max_{D_1}\mathcal{L}_{VAE_1} + \mathcal{L}_{GAN_1} + \lambda_{fm} \mathcal{L}_{FM}(x_1, \bar{x}_1), \label{eqn:VAEGAN1}
\end{align}
where $\lambda_{fm}$ is the loss weight and is set as $10$.
The loss function for the second domain can be similarly derived.
For training feature translation model, the objective function is,
\begin{align}
\min_{T_{1\to 2}} \max_{D_T} \mathcal{L}_{adv_1} + \lambda_f \mathcal{L}_{T_{1\to 2}}
+ \lambda_{fm} \mathcal{L}_{FM}(p_2, \bar{p}_2),
\end{align}
where $\bar{p}_2 = G_2(T_{1\to 2}(E_1(p_1)))$, 
$\lambda_f$ and $\lambda_{fm}$ are the loss weights that are set as
$60$ and $10$ respectively.

One intuitive training strategy is the joint optimization of the VAE-GAN and the feature translation network.
This is doable and seems to produce better results.
However, considering the case of multiple domains, say $N$ domains,
in order to provide image translation for any two different domains,
joint optimization needs to train $A_N^2$ ($= N\times (N-1)$) VAE-GANs,
whereas independent training in our strategy only needs to train $N$ VAE-GANs.
Moreover, we empirically find that joint training performs similarly to two-stage training.
Hence, we adopt this two-stage training procedure in the following experiments.

The feature translation network that we employ is $5$ fully connected layers for the task of profile to frontal faces
and is $9$ residual blocks for other tasks.
We train the VAE-GAN for $200$ epochs, and the feature translation network for $400$ epochs.
We both use the Adam solver with a learning rate of $0.0001$ and momentum parameters $\beta_1=0.5, \beta_2 = 0.999$, and the batchsize is set to $80$.
The training for feature translation network takes half an hour for face frontalization
and at most eight hours for other tasks using 4 Tesla P100 GPUs.
More details about the implementation are provided in the supplementary material.

\vspace{0.1cm}
\noindent\textbf{Inference Procedure.} We use the encoder $E_1$, the feature translation network $T_{1 \to 2}$ and the decoder $G_2$ to transform the image from the first domain to the second domain. For an input image $x_1$ in the first domain, the output image is $\bar{x}_2=G_2(T_{1 \to 2}(E_1(x_1)))$,
and the one-to-many translation results can be achieved by $G_2(T_{1 \to 2}(E_1(x_1)+\eta))$
with $\eta$ randomly sampled from a multi-variate Gaussian distribution: $\eta \sim \mathcal{N}(0,I)$.

\section{Experiments}
In this section, we first provide a thorough empirical analysis about our approach.
We then
present the comparison results of our framework on two challenging tasks:
labels-to-photo on the Cityscapes dataset\cite{cordts2016cityscapes} and face frontalization,
showing the superiority of our method.
Finally, we show more results to validate the generality of our proposed framework over a variety of tasks.

\subsection{Empirical Analysis}

We use the Cityscapes dataset for ablation experiments.
This dataset has $2975$ pairs of a semantic label map and a corresponding scene photo
for training and $500$ validation pairs for testing.
The unpaired data consists of $20000$ scene photos and $14000$ semantic label maps\footnote{These semantic label maps are obtained using the state-of-the-art scene segmentation algorithm~\cite{zhu2018improving} on the unpaired scene photos.}.
We adopt FCN-score, the same evaluation protocol as in Pix2Pix~\cite{isola2017image},
for quantitative comparison. 
We use the state-of-the-art segmentation network~\cite{zhu2018improving} 
(choose the same network with that outputting semantic label maps is fine since the outputted semantic label maps are considered as unpaired data) to evaluate the generated photos by the classification accuracy
against the semantic label maps from which the photos are synthesized. 
The groundtruth results are evaluated by applying the segmentation network over the real photos.

\begin{figure}[t]
	\begin{center}
		\includegraphics[width=0.48\linewidth]{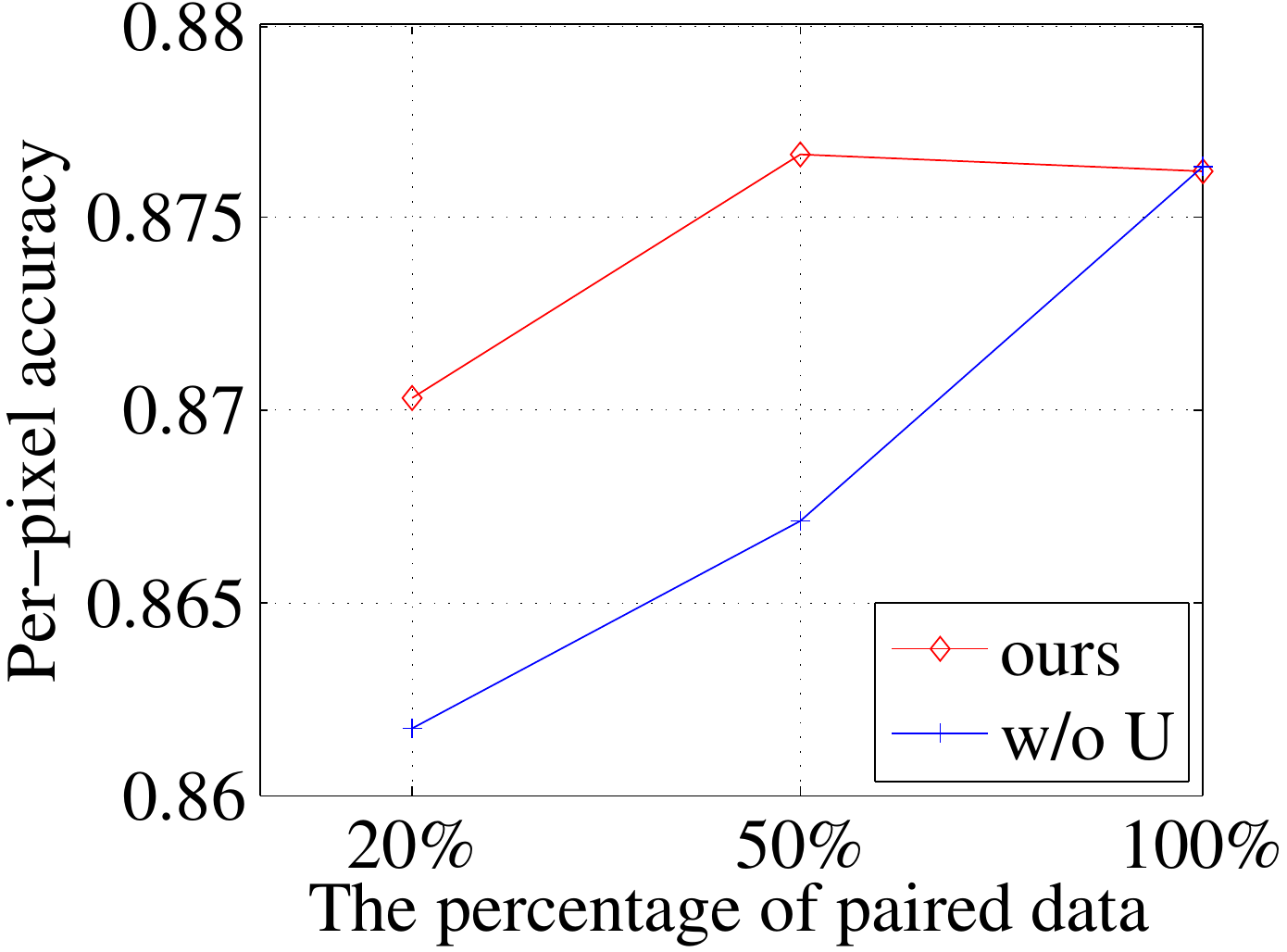}
		~~\includegraphics[width=0.48\linewidth]{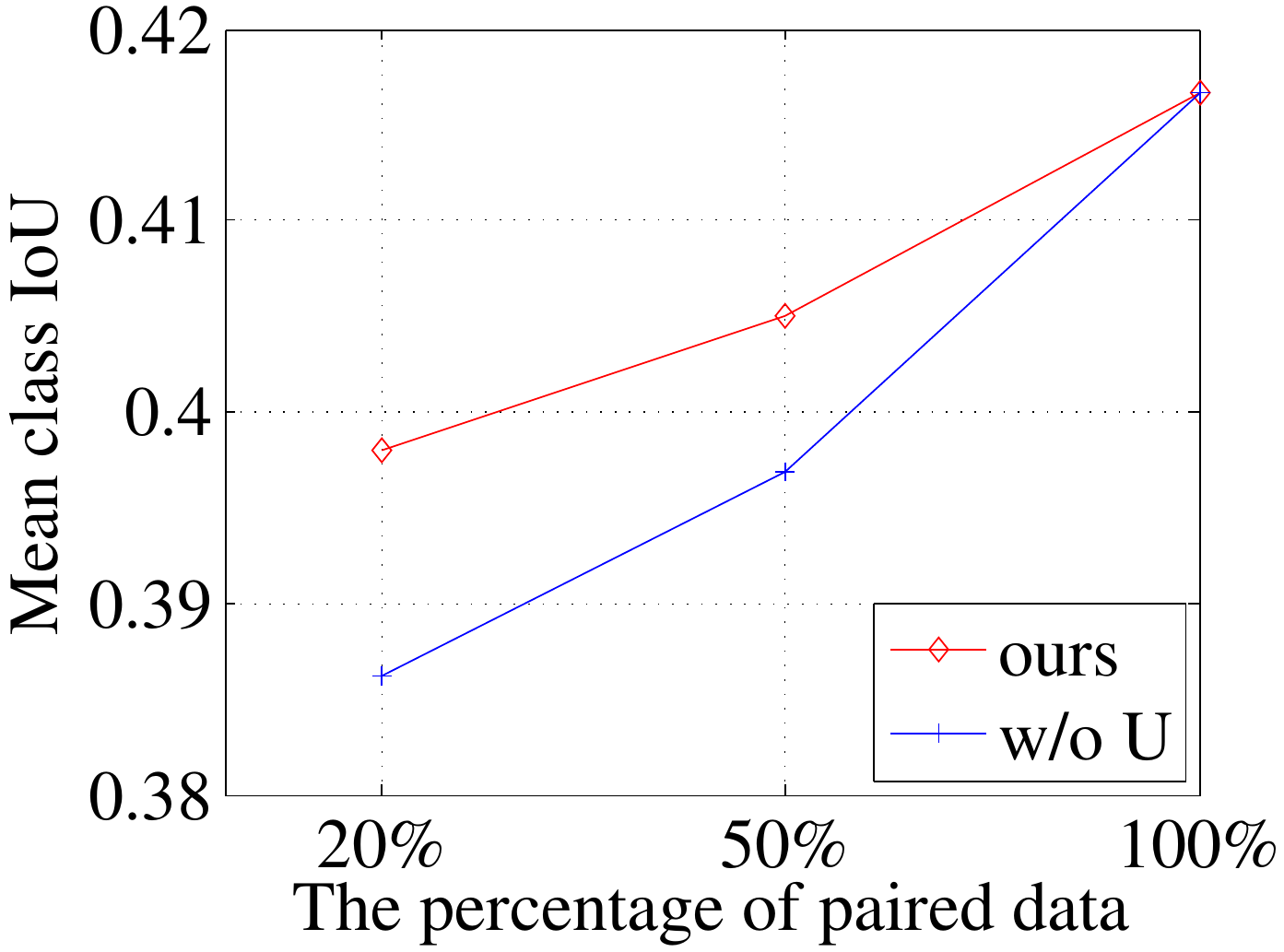}
	\end{center}
	\vspace{-0.2in}
	\caption{Illustrating the effect of unpaired data. \emph{w/o U} denotes our method without unpaired data.}
	\label{fig:data}
	\vspace{-0.6cm}
\end{figure}

\begin{table}[t]
	\centering
	\caption {Illustrating the effect of feature translation evaluated on Cityscapes labels$\rightarrow$photo. P (U) denotes paired (unpaired) data and IT (FT) denotes translation on image (feature) level.}
	\small
	\vspace{-0.35cm}
	\begin{tabular}{c|ccc}
		\hline
		Method& 	 Per-pixel acc. & Per-class acc. & Class IOU \\ 
		\hline
		IT+P	&  $ 0.856 $ & $ 0.445 $ & $ 0.356 $ \\
		IT+U\&P	 & $ 0.797 $ & $ 0.361 $ & $ 0.268 $ \\
		FT+U\&P (ours)	& $ \bf{0.876} $ & $ \bf{0.513} $ & $\bf{ 0.416 }$ \\
		\hline
	\end{tabular} 
	\label{tab:ablation}
	\vspace{-0.3cm}
\end{table}

\begin{table}[t]
	\centering
	\caption{Illustrating the effect of loss for feature transformation.}
	\vspace{-0.35cm}
	\footnotesize
	\setlength\tabcolsep{2pt} 
	\begin{tabular}{c|ccc}
		\hline
		Loss & Per-pixel acc. & Per-class acc. & Class IOU \\ 
		\hline
		$\mathcal{L}_{T_{1\to 2}}$ & $ 0.834 $ & $ 0.431 $ & $ 0.336 $ \\
		$\mathcal{L}_{adv_1}(z_2,\bar{z}_2)+ \lambda_f \mathcal{L}_{T_{1\to 2}}$ & $ 0.851 $ & $ 0.465 $ & $ 0.364 $ \\
		Equation~(\ref{eqn:translation}) (ours) & $ \bf{ 0.876 }$ & $\bf{ 0.513} $ & $ \bf{0.416} $\\
		\hline
	\end{tabular}
	\label{tab:loss_feature_transformation}
	\vspace{-0.6cm}
\end{table}

\begin{figure*}[t]
    \centering
    \includegraphics[width=1.0\linewidth]{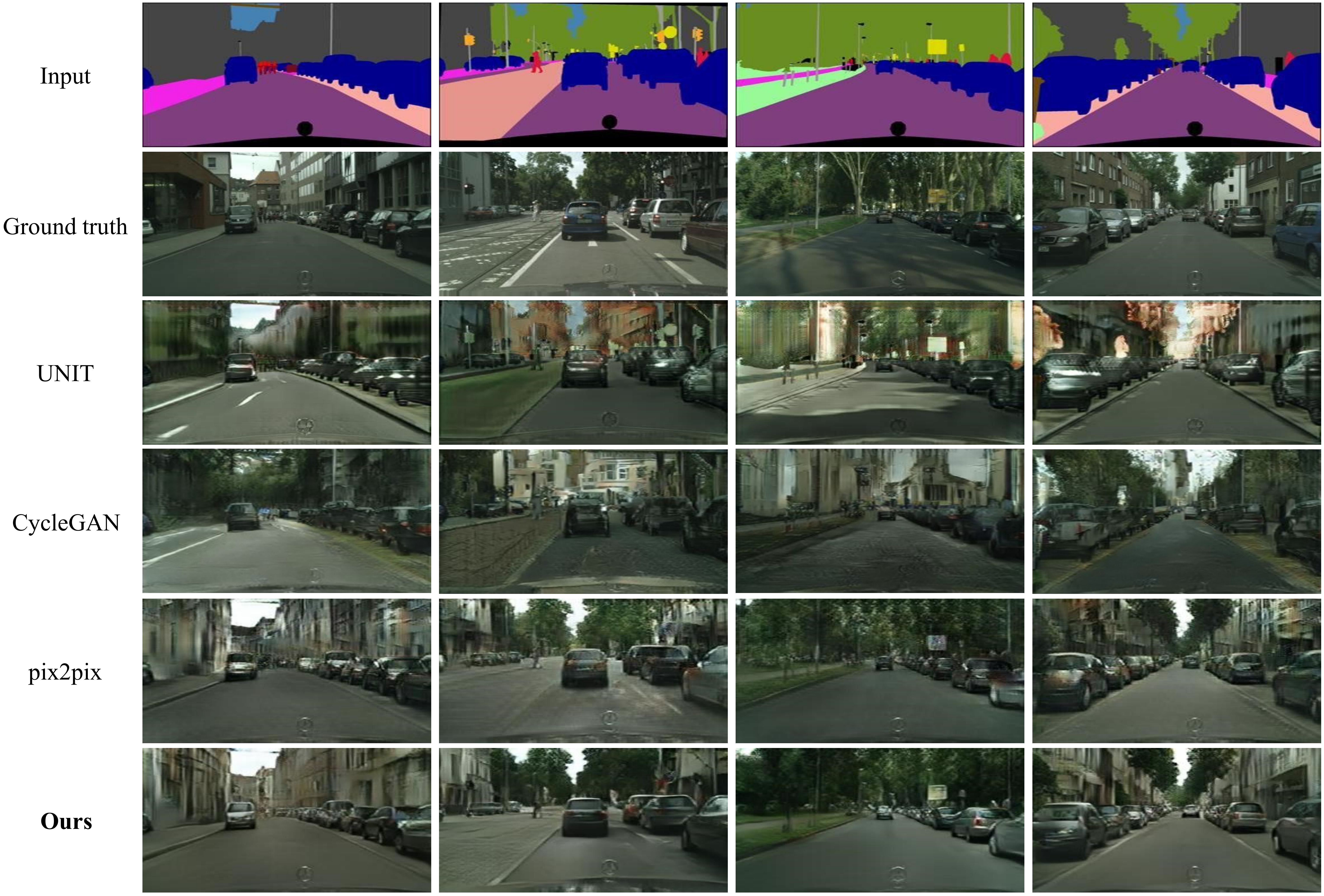}
    \vspace{-0.6cm}
    \caption{Comparison results of semantic labels to scene photos on the Cityscapes dataset for different methods. From top to bottom: input, ground truth, UNIT~\cite{liu2017unsupervised} CycleGAN~\cite{zhu2017unpaired}, pix2pix~\cite{isola2017image}, and ours.}
    \label{fig:comparison_to_baselines}
    \vspace{-0.5cm}
\end{figure*}

\vspace{0.1cm}
\noindent
{\bf Effect of Unpaired Data.}
Supervision is of great value for many computer vision tasks, including image-to-image translation. 
When there exists sufficient supervision,
our framework also achieves good performance using only provided paired data.
However, large scale supervised data is hard to obtain, requiring a lot of labor.
In this case,
the easy-acquired unsupervised data
can be used in our approach to train the VAE-GAN for getting informative
latent features.
To show this,
we study the results of our model and \emph{w/o U} (our framework without unpaired data) over different sizes of paired data. 
The comparison is shown in Figure~\ref{fig:data}.
We observe that the improvement brought by unpaired data is more significant as the size of data pairs becomes smaller,
which is desirable since many real world tasks have only minimal supervision.
When using all the paired data, the results with additional unpaired data show little improvement.
The reason may be that the supervision information is dominant in this case.
%We also plot the results of our-U-F in the figure,
%and the comparison with our-U shows 
%that feature translation achieves consistent improvement.

\vspace{0.1cm}
\noindent
{\bf Effect of Feature Translation.}
To show the importance of feature translation, we study two variants of our method:
1) translation on the image level, denoted as \emph{IT+U\&P}, for which we adopt the idea of UNIT on unpaired data and the idea of pix2pix on paired data;
2) translation on the image level with only paired data, denoted as \emph{IT+P}, for which we adopt the idea of pix2pix.
All the models are trained under the same setting.
The comparison is given in Table~\ref{tab:ablation}.
Our model achieves large improvement compared with \emph{IT+U\&P}, verifying the 
effectiveness of feature translation.
Comparing \emph{IT+U\&P} with \emph{IT+P}, it can be seen that
adding additional unpaired data does not get improvement and instead cause performance drop.
This indicates that the unpaired data should be exploited properly in order to improve the performance.

\begin{table}[t]
	\centering
	\caption {Quantitative comparison in terms of FCN-scores for different methods on Cityscapes labels$\rightarrow$photo.
		Pix2pix* denotes the results of pix2pix using our network structure.
		Our approach achieves the best performance. In addition, our method using only $20\%$
		($\approx 600$) pairs also performs better than the baselines.}
	\vspace{-0.35cm}
	\small
	\setlength\tabcolsep{3pt} 
	\begin{tabular}{lcccc}
		\hline
		Loss & Per-pixel acc. & Per-class acc. & Class IOU \\ \hline
		CycleGAN~\cite{zhu2017unpaired} & $ 0.509 $ & $ 0.231 $ & $ 0.152 $ \\
		UNIT~\cite{liu2017unsupervised} & $ 0.658 $ & $ 0.240 $ & $ 0.189 $ \\
		pix2pix~\cite{isola2017image} & $ 0.826 $ & $ 0.470 $ & $ 0.354 $ \\
		pix2pix*  & $ 0.856 $ & $ 0.445 $ & $ 0.356 $ \\
		pix2pix* ($20\%$ pairs) & $ 0.747 $ & $ 0.358 $ & $ 0.264 $ \\
		\hline
		ours ($20\%$ pairs) & $ 0.870 $ & $ 0.513 $ & $ 0.398 $ \\
		ours & $ \bf{0.876} $ & $\bf{ 0.513 }$ & $ \bf{ 0.416 }$ \\ \hline
		Ground truth & $ 0.897 $ & $ 0.615 $ & $ 0.506 $ \\
		\hline
	\end{tabular} 
	\label{tab:comparison_with_other_methods}
	\vspace{-0.1cm}
\end{table}

\vspace{0.1cm}
\noindent
{\bf Effect of Loss for Feature Transformation.} 
The loss that guides the feature transformation can be other two alternatives: 1) only using $L_1$ loss over features, namely $\mathcal{L}_{T_{1\to 2}}$; 2) imposing the adversarial loss on the features instead of the images, denoted as  $\mathcal{L}_{adv_1}(z_2,\bar{z}_2)+ \lambda_f \mathcal{L}_{T_{1\to 2}}$,
where $\mathcal{L}_{adv_1}(z_2,\bar{z}_2) = \mathbb{E}[D_T(z_2)^2] + \mathbb{E}[(1-D_T(\bar{z}_2))^2]$ and $\bar{z}_2 = T_{1\to 2}(z_1)$. 
We present the comparison results in Table~\ref{tab:loss_feature_transformation},
showing that our loss achieves the best performance.
We have two observations: 1) adding adversarial loss helps generate better results;
2) adversarial loss encouraging generate realistic synthesized images is more effective than encouraging the translated features to be faithful.

\subsection{Results}

\begin{figure}[t]
    \centering
    \includegraphics[width=0.95\linewidth]{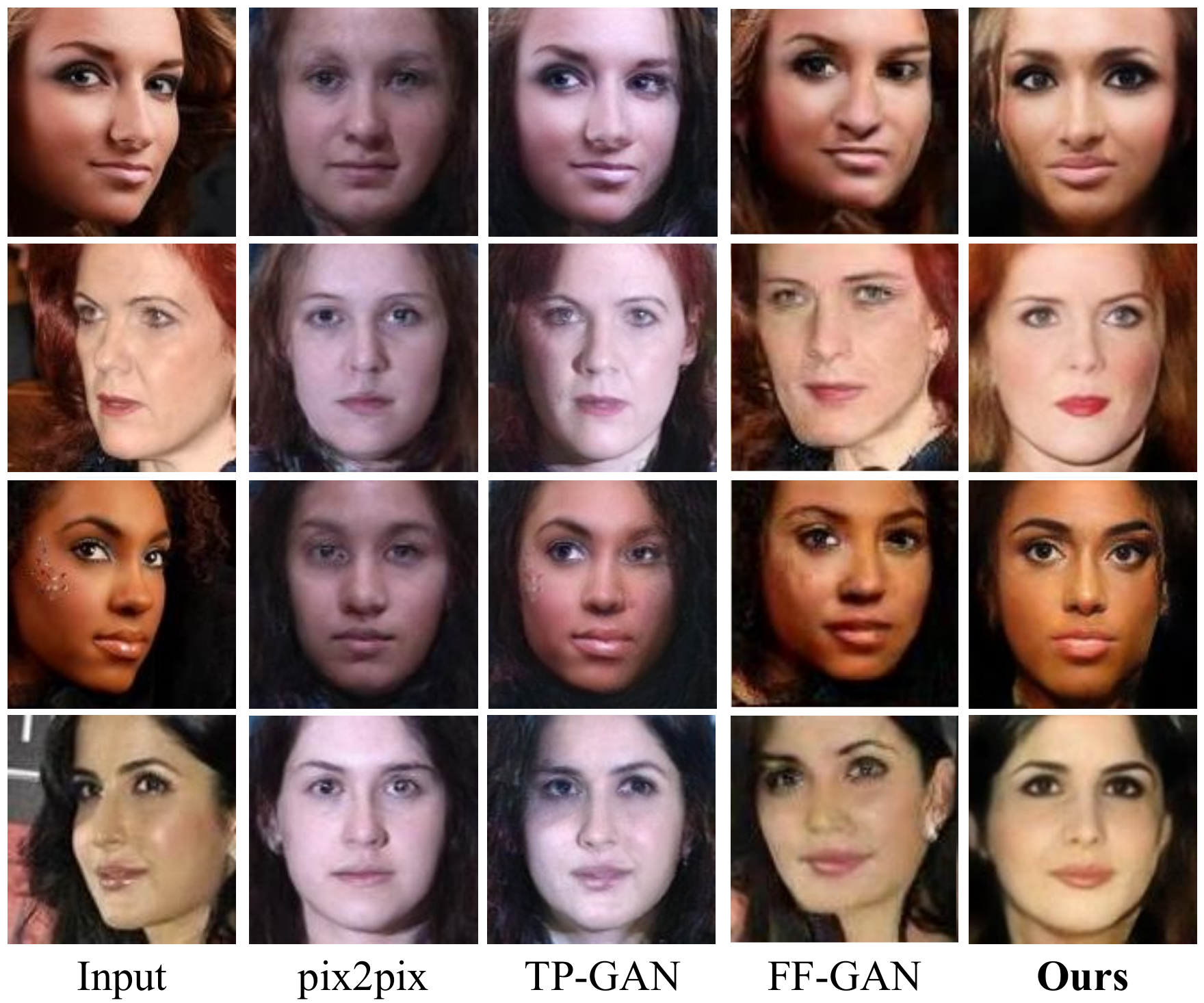}
    \vspace{-0.2cm}
    \caption{Visual comparison of face frontalization for different methods on the IJB-A and AFLW2000 dataset.}
    \label{fig:face_frontalization}
    \vspace{-0.6cm}
\end{figure}

\subsubsection{Results of Semantic Labels to Street Scenes}
We show the results on the task of semantic labels to photos on the Cityscapes dataset,
which has been widely studied in previous works.
We compare our framework with three state-of-the-art baselines:
pix2pix~\cite{isola2017image}, an image-to-image translation work which uses the adversarial loss and paired data;
CycleGAN~\cite{zhu2017unpaired}, an unsupervised image-to-image translation work which introduces a cycle-consistency loss;
UNIT~\cite{liu2017unsupervised}, also an unsupervised image-to-image translation work based on the shared latent space assumption.

The visual comparison is shown in Figure~\ref{fig:comparison_to_baselines}. It can be easily seen that our generated results are more clear with less artifacts.
The numerical comparison in terms of FCN-scores is shown in 
Table~\ref{tab:comparison_with_other_methods}. 
As expected, the supervised method pix2pix~\cite{isola2017image} performs better than the unsupervised methods CycleGAN~\cite{zhu2017unpaired} and UNIT~\cite{liu2017unsupervised}.
For fair comparison,
we also present the results of pix2pix which adopt the same network architecture with ours,
denoted as pix2pix*.
Among all the compared methods, 
our framework achieves the best scores that are close to the groundtruth.
It is worth noting that our method using only $20\%$ pairs performs better than pix2pix with full supervision, further indicating the powerfulness of the proposed feature translation.

\begin{figure}[t]
    \centering
    \includegraphics[width=0.95\linewidth]{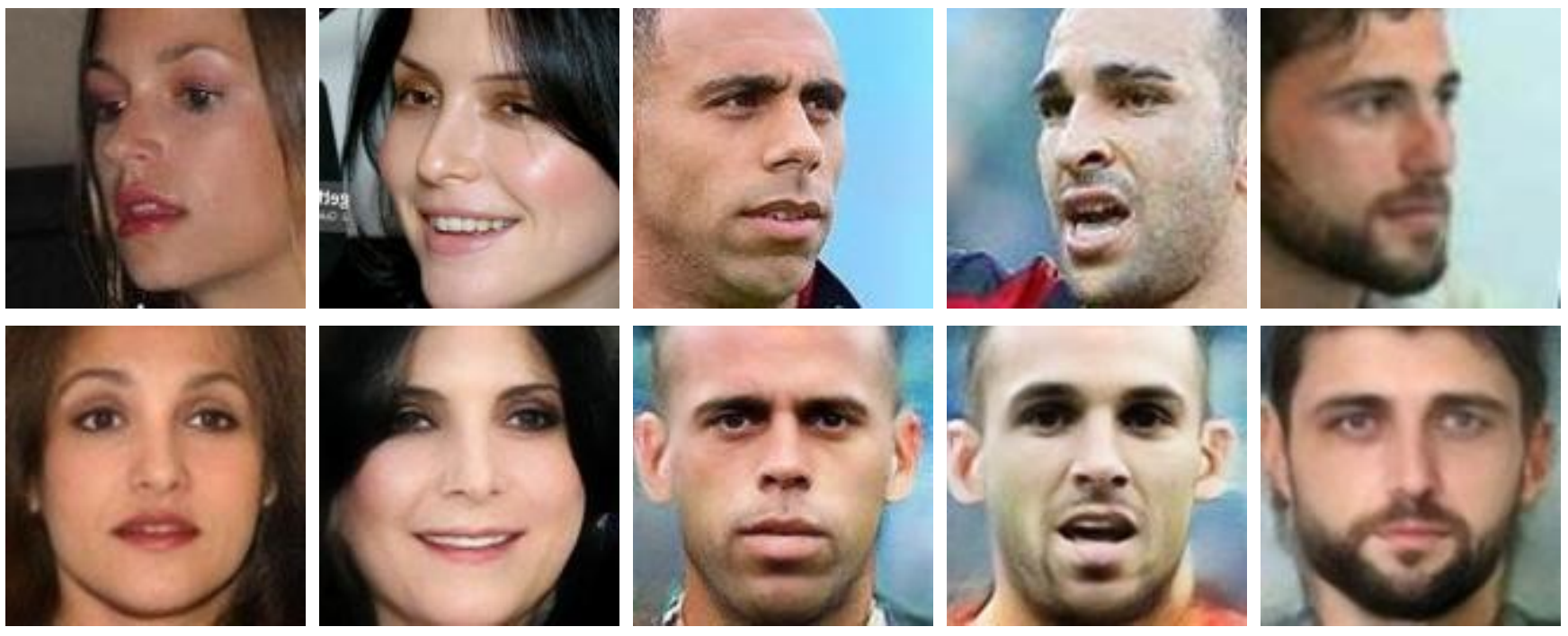}
    \vspace{-0.2cm}
    \caption{Face frontalization results on VGGFace2 dataset.}
    \label{fig:face_frontalization_vggface2}
    \vspace{-0.2cm}
\end{figure}

\begin{figure}[t]
    \centering
    \includegraphics[width=0.95\linewidth]{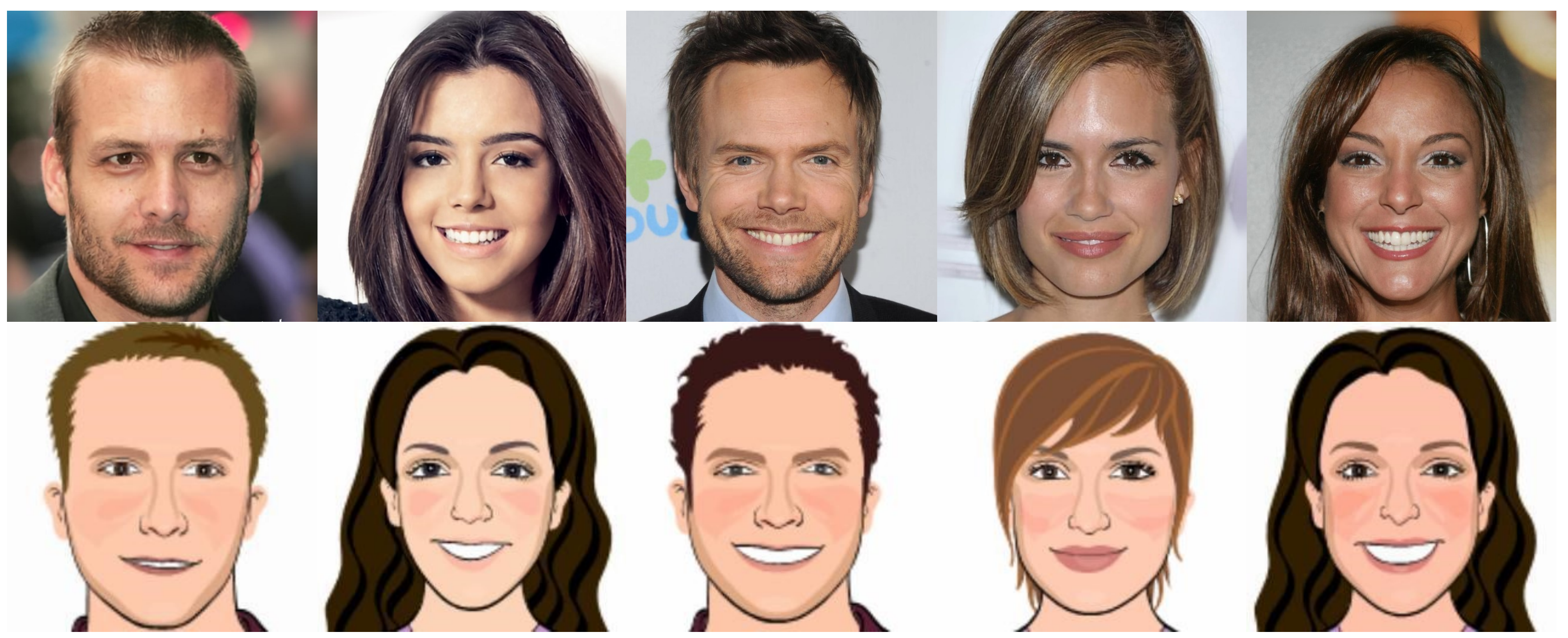}
    \vspace{-0.2cm}
    \caption{Example training pairs of our face$\rightarrow$cartoon dataset.}
    \label{fig:face_to_cartoon_examples}
    \vspace{-0.6cm}
\end{figure}

\begin{figure*}[t]
    \centering
    \includegraphics[width=1.0\linewidth]{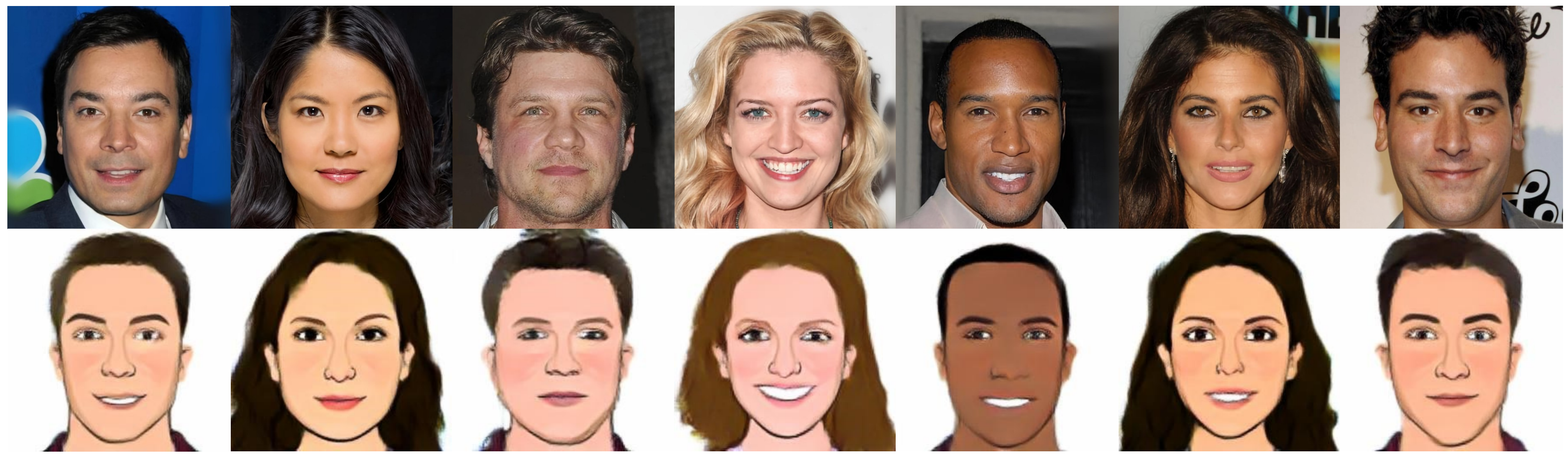}
    \vspace{-0.6cm}
    \caption{Synthesized results of real faces to cartoons. The first row is the input faces, the second row is the generated results.}
    \label{fig:face_to_cartoon}
\end{figure*}

\begin{figure*}[t]
	\centering
	\includegraphics[width=1.0\linewidth]{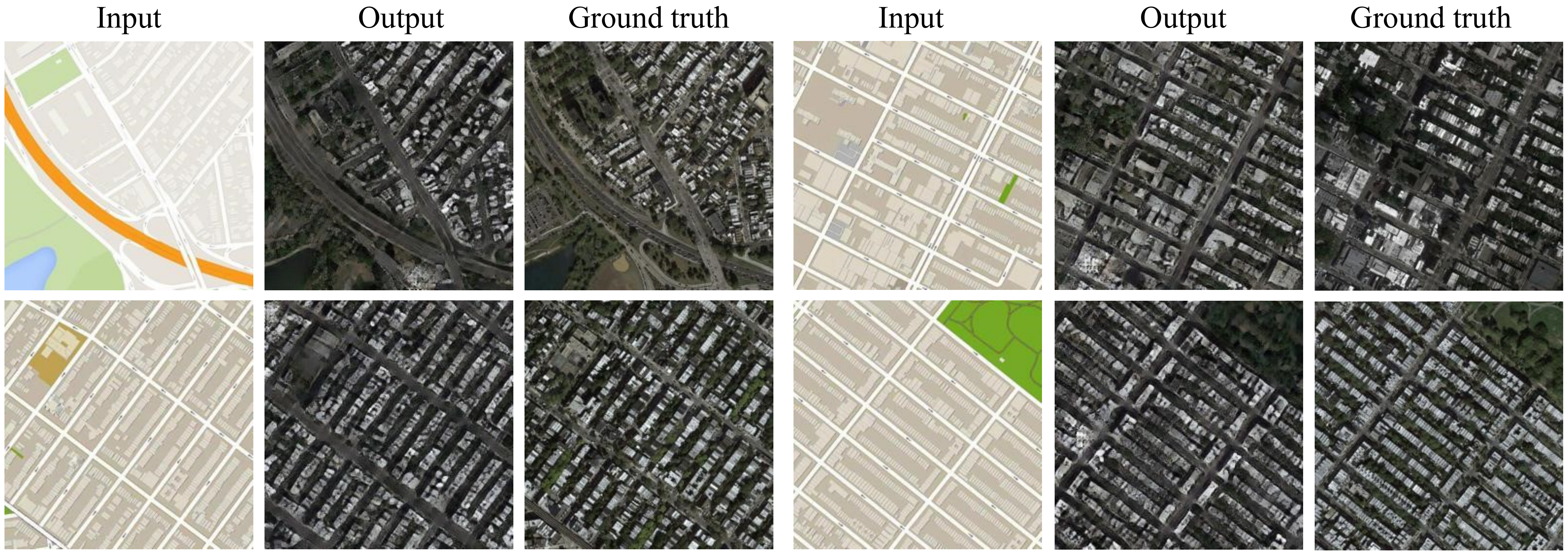}
	\vspace{-0.6cm}
	\caption{Example results of our method on Google map$\rightarrow$photo, compared to ground truth.}
	\label{fig:map_to_aerial}
\end{figure*}

\begin{figure*}[t]
	\centering
	\includegraphics[width=1.0\linewidth]{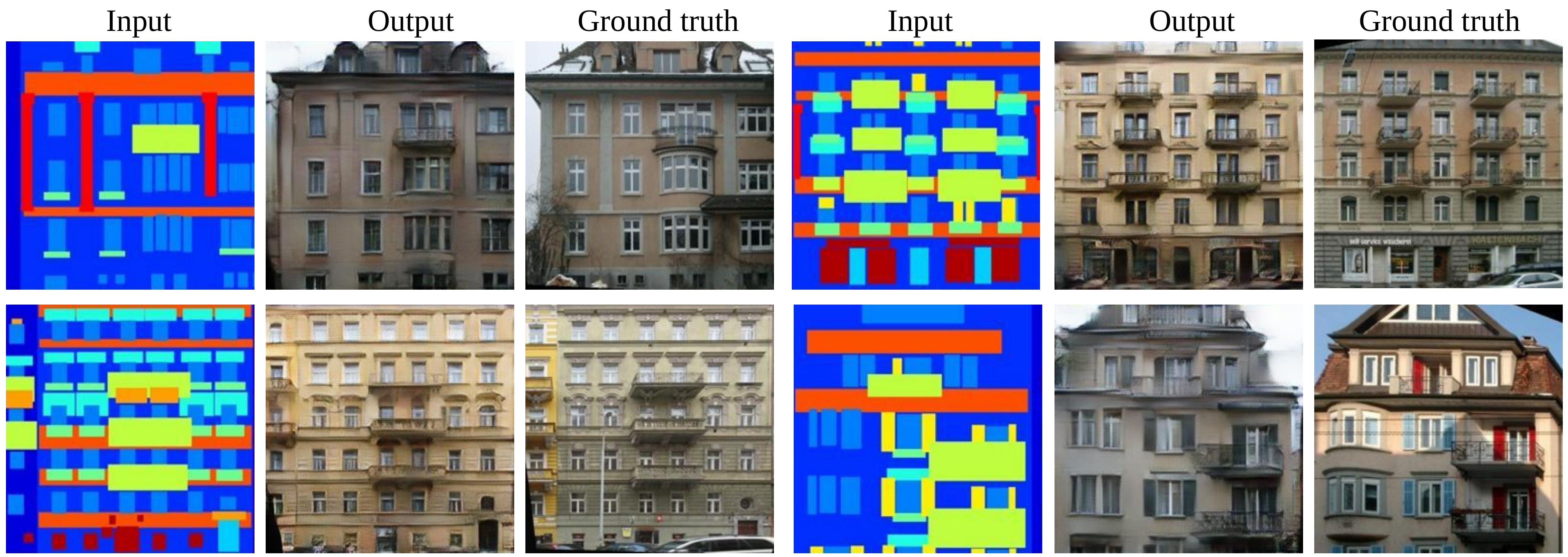}
	\vspace{-0.6cm}
	\caption{Example results of our method on architecture labels $\rightarrow$photos, compared to ground truth.}
	\label{fig:architecture_to_photos}
\end{figure*}

\begin{figure*}[t]
	\centering
	\includegraphics[width=1.0\linewidth]{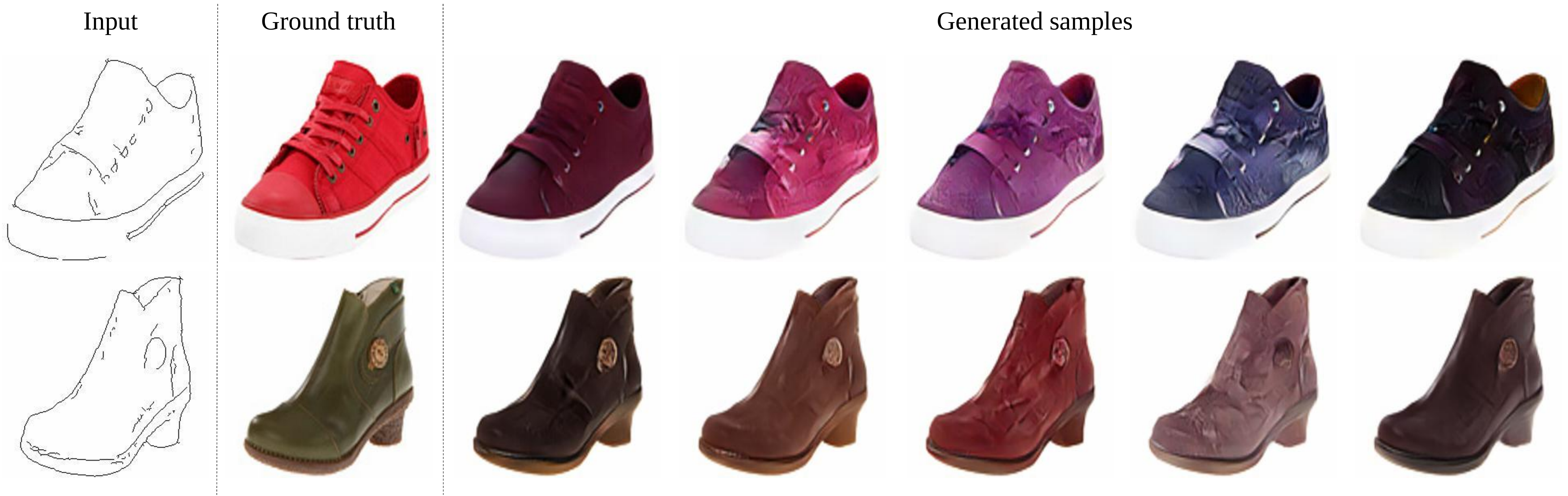}
	\vspace{-0.6cm}
	\caption{Example results of our method on edges$\rightarrow$shoes, The left column shows the input. The second column shows the ground truth. The last five columns show the randomly generated samples of our approach.}
	\label{fig:edge_to_shoes}
\end{figure*}

\begin{figure*}[h!]
	\centering
	\includegraphics[width=1.0\linewidth]{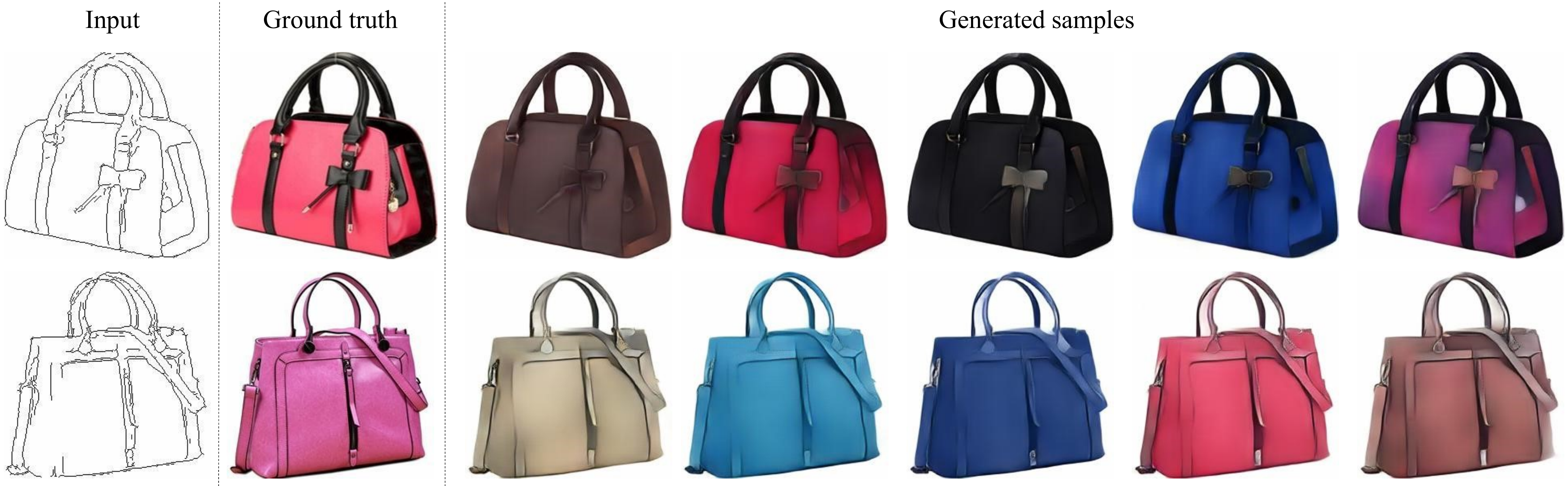}
	\vspace{-0.6cm}
	\caption{Example results of our method on edges$\rightarrow$bags, The left column shows the input. The second column shows the ground truth. The last five columns show the randomly generated samples of our approach.}
	\label{fig:edge_to_bags}
\end{figure*}

\subsubsection{Results of Face Frontalization}
Face frontalization is the process of generating frontalized faces given single unconstrained photos
with large head pose variations.
In fact, it
can be treated as an image-to-image translation task: profile-to-front, where paired data is extremely hard to get. 
Compared to other cross-domain image generations tasks,
face frontalization is more difficult due to significant geometric deformation.

We show that our framework is capable of handling image-to-image translation with large shape gaps.
Experiments are conducted on the widely used datasets: VGGFace2\cite{Cao18} containing unpaired images from 
$9000$ identities and Multi-PIE\cite{gross2010multi} containing paired images from $337$ identities.
In order to preserve the personality after translation, we add an identity preservation loss,
which is the widely used feature matching loss over the identity features of the input face and the frontalized output.

We compare our results with two state-of-the-art face frontalization techniques, TP-GAN\cite{huang2017beyond}
and FF-GAN\cite{yin2017towards}, as well as one representative image translation method, pix2pix~\cite{isola2017image}.
TP-GAN\cite{huang2017beyond} uses a global and multiple local aware GAN architectures to learn the translation using paired data. FF-GAN\cite{yin2017towards} combines both the 3D morphable model and the GAN structure to generate frontal viewed faces. 
Figure~\ref{fig:face_frontalization} provides the visual comparison results, where we choose the input samples appeared in TP-GAN and FF-GAN.
We can see that the generated results 
of pix2pix and TP-GAN fail to keep the skin color of the input face which causes uncertainty about the identity, although
TP-GAN captures the details of the face because of the multiple local network paths.
While FF-GAN generates near-frontal and blurry faces. 
In contrast, our model produces photo-realistic and reasonable results.
More results of our model for face frontalization are shown in Figure~\ref{fig:face_frontalization_vggface2}.

%\subsubsection{Results of face-to-cartoon}
%We further apply our method to a main application challenge, cartoon generation given a facial image. 
%The Celeba-HQ face dataset in PGGAN\cite{karras2017progressive} is used to compose the images from the face domain.
%We use a pose estimator to remove the faces with large pose, and finally get about $10000$ face images. 
%To collect images from the cartoon domain,
%we use a tool from \cite{wen2013cartoon} and get about $1200$ paired data for face-to-cartoon translation. 
%Some example pairs are shown in Figure~\ref{fig:cartoonexamples}.
%We randomly sample ? pairs as the supervised information for training and use the rest for testing.
%For unpaired images from cartoon domain, we also use that tool to randomly generate about $18000$ face cartoon images.
%? residual blocks.
%
%Figure \ref{fig:face_to_cartoon} shows the results of face-to-cartoon generation. We can observe that 

\subsubsection{More Applications}
We further show more applications to demonstrate the generality of our proposed framework:

\begin{itemize}
\itemsep-0.05in
\item \emph{face$\rightarrow$cartoon}: The Celeba-HQ face dataset~\cite{karras2017progressive} is used to compose the images of the face domain. We use a pose estimator to remove the faces with large pose, and finally get $10000$ face images. To collect images of the cartoon domain, we use an interactive tool from~\cite{wen2013cartoon} and manually labeled $1200$ paired data for face-to-cartoon translation. Some example pairs are illustrated in Figure~\ref{fig:face_to_cartoon_examples}.
\item \emph{Map$\rightarrow$aerial photo}: trained on the data scraped from Google Maps.
\item \emph{Architectural labels$\rightarrow$photo}: trained on the CMP Facades \cite{tylevcek2013spatial}.
\item \emph{Edges$\rightarrow$photo}: trained on data from \cite{zhu2016generative} and \cite{yu2014fine}; the binary edge maps are generated using the HED edge detector \cite{xie2015holistically} plus postprocessing.
%\item \emph{Sketch$\rightarrow$photo}: tests edges$\rightarrow$photo models on human-drawn sketches from \cite{eitz2012humans}.
\end{itemize}

Implementation details on each dataset are presented in the supplementary material. 
The qualitative results are given in Figure~\ref{fig:face_to_cartoon},~\ref{fig:map_to_aerial},~\ref{fig:architecture_to_photos},~\ref{fig:edge_to_shoes},~\ref{fig:edge_to_bags}. 
We show that our approach is able to achieve one-to-many translations in Figure~\ref{fig:edge_to_shoes} and Figure~\ref{fig:edge_to_bags}.

%\section{Discussions}
%\noindent
%{\bf The choice of two-stage training.}
%Our framework adopts a two-stage training strategy: first train a VAE-GAN for each domain,
%and then train the translation network with fixed VAE-GAN.
%One intuitive possible training strategy is the joint optimization of the VAE-GAN and the translation network.
%We empirically found that such joint training performs similar with the two-stage training.
%On the other hand,
%considering there are multiple domains, \eg, N domains,
%$A_N^2$ ($= N\times (N-1)$) VAE-GANs and $A_N^2$ translations are needed in order to provide image translation for any two different domains if joint training is used.
%In contrast, only $N$ VAE-GANs and $A_N^2$ translations are needed for two-stage training.
%This motivates us to prefer the two-stage training.
%
%
%
%
%\noindent
%{\bf Extension to unsupervised setting.}
%In the case of unsupervised setting that no paired images is available,
%our approach can be easily generalized to that case by combining CycleGAN~\cite{zhu2017unpaired} with feature translation.  
%Note that this extension is different from the feature cycle-consistency in previous works where there is an image generated through that cycle. 
%In particular, we apply CycleGAN training over the two independent sets of features $\mathcal{F}_u = E_u (\mathcal{X}_u)$ and 
%$\mathcal{F}_v = E_v (\mathcal{X}_v)$ 
%to simultaneously learn the feature translation models $T_{u\to v}$ and $T_{v\to u}$. We show an example comparison
%over ? dataset, and compared with UNIT? DTN?

\section{Conclusion}
In this paper, we propose a novel framework for semi-supervised image-to-image translation. 
The novelty mainly stems from the idea of feature translation, which translates features instead of images, resulting in improved quality and stability of image translation.
Experiments on
a wide variety of applications verify the effectiveness of the proposed method.
% In the future, we would try to combine CycleGAN with feature translation. We would apply cycle-consistency training over the two independent sets of features and simultaneously learn two feature translation networks. We can also try one-to-many image translation with the help of the latent feature space.

%\clearpage

{\small
\bibliographystyle{ieee}
\bibliography{im2im}
}

\end{document}